\documentclass{svjour3}                     
\smartqed  
\usepackage{amsmath, amssymb}
\usepackage[ruled,linesnumbered,vlined]{algorithm2e}
\usepackage{enumitem}

\newtheorem{defi}{Definition}

\makeatletter
\newcommand\Envelope{{\fontencoding{U}\fontfamily{ding}\selectfont\symbol{'014}}}
\makeatother
\usepackage{graphicx}

\journalname{MISTA 2019}
\begin{document}

\title{Robustness Approaches for the Examination Timetabling Problem under Data Uncertainty\thanks{Research funded in parts by the School of
Engineering of the University of Erlangen-Nuremberg.}}

\author{Bernd Bassimir         \and
  Rolf Wanka
}

\institute{Bernd Bassimir (\raisebox{-1.5pt}{\Envelope})\at
  Department of Computer Science,
  University of Erlangen-Nueremberg, Germany\\
  \email{bernd.bassimir@fau.de}           
  \and
  Rolf Wanka \at
  Department of Computer Science,
  University of Erlangen-Nueremberg, Germany\\
  \email{rolf.wanka@fau.de}           
}

\maketitle

\begin{abstract}
  In the literature the examination timetabling problem (ETTP) is
  often considered a post-enrollment problem (PE-ETTP). In the real
  world, universities often schedule their exams before students
  register using information from previous terms. A direct consequence
  of this approach is the uncertainty present in the resulting
  models. In this work we discuss several approaches available in the
  robust optimization literature. We consider the implications of each
  approach in respect to the examination timetabling problem and
  present how the most favorable approaches can be applied to the
  ETTP. Afterwards we analyze the impact of some possible
  implementations of the given robustness approaches on two real world
  instances and several random instances generated by our instance
  generation framework which we introduce in this work.
\end{abstract}

\section{Introduction}
\label{sec:intro}

In every academic term,
universities are faced with a number of different aspects of
academic timetabling. 
Academic timetabling is divided into two distinct,
but similar problems, namely the \emph{Course Timetabling Problem} (CTTP)
(for a comprehensive overview, see M\"uhlenthaler's
monograph~\cite{M:15}) prior to each term
and the \emph{Examination Timetabling Problem} (ETTP) at the end of the term.

CTTP is the task to assign one timeslot and one room to each lecture
that is held in this term, creating a timetable that can be repeated
each week over the complete term.  Besides the problem of checking
whether there is an admissible timetable at all, there is also an
optimization variant, if there is a way to assess timetables and look
for ``good'' (or best possible) timetables.

In the ETTP, 
one timeslot and one or more rooms must be assigned to each exam.  An
exam is held only once per term and as such the timetable, in contrast
to the CTTP, needs not to be repeatable.  At many universities, the
exams are held at the end of the term in a period
of a few weeks after the end of the lectures.  Again, there is also an
optimization variant. For a detailed model of the ETTP see
\cite{McCollum2012}.

Both mentioned problems are NP(O)-complete
and, hence, no polynomial time (exact) algorithms are known. 
As each university is faced with the ETTP and the CTTP
several times per year,
many different approaches are proposed
in the literature to solve these problems automatically.

In the literature the ETTP is often described as a problem where the
number of students that are attending an exam are given as an input to
the scheduling algorithm, which is sometimes called
\emph{post-enrollment} ETTP or is at least treated as a
\emph{post-enrollment} ETTP, e.\,g., see
\cite{Eley:2006,Gogos2012,ITC-M}. First the students register for
their different exams and after registration is finished the
optimization takes place. In this approach all parameters are known,
as we have the exact number of students registered for an exam and
exams are in conflict, i.\,e., should not be scheduled at the same
time, if there are students taking both exams. In practice however
some universities generate their schedules for the exams before this
registrations takes place to provide a time and date for the exams
when students register, e.\,g., see
\cite{BW:18,Cataldo2017,Demeester2012}. Most often the values for the
number of students per exam are taken from the last year as well as
the conflicts that were present in the previous year.

In this work we consider the number of students attending an exam as
well as the number of students inducing a conflict for a pair of exams
to be random variables.  Given that these variables are not fixed as
in the classical post-enrollment model we have to account for the
uncertainty and consequently the deviation from the estimated values
of the actual values when students eventually register for their
exams. Instead of improving the estimations, which might not be
possible, we focus on handling the uncertainty in the optimization process. To
handle this uncertainty in the optimization process we discuss several
approaches of robust optimization and their application on the
ETTP. We furthermore introduce a new instance generation framework,
from which we obtain a more diverse test set.

In Sect.~\ref{sec:model} we introduce the basic model used in this
work and the implications of the uncertainty on this model. In
Sect.~\ref{sec:robust} we introduce several different strategies for
robust and stochastic optimization to deal with this uncertainty and
in Sect.~\ref{sec:framework} we introduce a framework for the
generation of random instances and the simulated annealing algorithm
we used in our experiments. In Sect.~\ref{sec:experiments} we compared
implementations of two robustness approaches on two real world
instances taken from the University of Erlangen-N\"urnberg and random
instances generated with the newly introduced instance generation
framework.

\section{Model}
\label{sec:model}

In~\cite{McCollum2012}, McCollum et\,al. introduce an extensive model
for the Examination Timetabling Problem. For the purpose of this paper
we use a reduced version of this model with only a subset of the soft constraints.

\begin{defi}
  An instance of the Examination Timetabling Problem is represented as follows.
  \begin{itemize}
  \item $\cal E$: A set of exams
  \item $\cal R$: A set of rooms
  \item $\kappa: {\cal R}\rightarrow \mathbb{N}$: the capacity
    available for each room and
    $\kappa: {\cal P}({\cal R}) \rightarrow \mathbb{N}$ as the
    canonical extension such that for
    $X\subseteq {\cal R}: \kappa(X):= \sum_{r \in X} \kappa(r)$
  \item $\nu \in \mathbb{N}^{|\cal E|}$: A vector specifying for each
    exam how many students are attending
  \item A conflict matrix $C \in \mathbb{N}^{|\cal E|\times |E|}$
    specifying the number of overlapping students of two exams
  \item A set of soft constraints with associated weights
  \end{itemize}
\end{defi}
\paragraph{Hard Constraints}

To be feasible, a timetable must meet the following hard constraints:
\begin{enumerate}[label=(H\arabic*),labelindent=3pt,leftmargin=*]
\item each exam is assigned to exactly one timeslot \label{enum:timeslot}
\item each exam is assigned to one or more rooms \label{enum:room}
\item two exams that are in conflict are not scheduled at the same time 
  \label{enum:conflicting}
\item no room is used at the same time by two exams \label{enum:room_conf}
\item the sum of capacities of the assigned rooms is larger than the number of students taking the exam \label{enum:cap}
\end{enumerate}
Most of this model is similar to the model in~\cite{McCollum2012}
except that we allow multiple rooms per exam.

\paragraph{Soft Constraints}
In our reduced model we use the soft constraints \emph{two-in-a-row},
\emph{two-in-a-day} and \emph{period-spread} as defined in
\cite{McCollum2012}. 
\begin{enumerate}[label=(S\arabic*),labelindent=3pt,leftmargin=*]
\item \emph{two-in-a-row}: If two exams are in conflict according to
  $C$ they should not be assigned to two adjacent timeslots on the
  same day. \label{enum:two-in-a-row}
\item \emph{two-in-a-day}: If two exams are in conflict according to
  $C$ they should not be assigned to two timeslots on the same
  day. Note that we exclude the directly adjacent timeslot to avoid
  double counting. \label{enum:two-in-a-day}
\item \emph{period-spread}: If two exams are in conflict according to $C$
  they should not be assigned to timeslots less than $\lambda$ apart. \label{enum:period-spread}
\end{enumerate}
Each violation of a soft criterion induces a penalty corresponding to
the number of students that are in conflict given by $C$.  Given a
feasible timetable, i.\,e., a timetable that satisfies hard
constraints \ref{enum:timeslot} - \ref{enum:cap}, we can formulate the
objective value as the weighted sum of the penalties induced by the
soft constraints \ref{enum:two-in-a-row} -
\ref{enum:period-spread}. As a violation of soft constraint
\ref{enum:two-in-a-row} infers a violation of soft constraint
\ref{enum:two-in-a-day} we exclude the adjacent timeslot in the
calculation of the violations of soft criterion
\ref{enum:two-in-a-day}. For the soft criterion
\ref{enum:period-spread} we ignore overlaps and compensate this in the
weight. The weights of the penalties can vary for each university and
therefore are not fixed in the model. For a possible selection see
Sect.~\ref{sec:experiments}.

\subsection{Examination timetabling under uncertainty}
\label{sec:uncertain_model}

In real world examination timetabling the schedule is sometimes generated
before students register for exams. This approach can be called
Curriculum-Based analogue to the course timetabling problem. In such a
situation the exact number of students attending an exam and the
number of students inducing conflicts for exams are not known at the
time of optimization. However usually we have static information
available provided by the curriculum model.  There is the curriculum
data ${\rm Curr}\!\!: {\cal E} \rightarrow {\cal P}({\cal M})$
assigning each exam a set of major and semester combinations
${\cal M}$, such that students in this combination of major and
semester can attend the respective exam and the registration data
${\rm Enroll}\!\!: {\cal M} \rightarrow \mathbb{N}$ specifying the
number of students registered for each major and semester combination.

We treat the number of students attending an exam $e$ as a random
variable $\nu_e$ with an unknown discrete distribution on the support
$\{1,2,\ldots, \sum_{c \in {\rm Curr}(e)} {\rm Enroll}(c)\}$. Note
that we exclude $0$ as this would imply that nobody would take the
exam and therefore the exam would not be scheduled. We ignore this
case as all exams have to be scheduled. Furthermore as there are
optional courses we can not quantify the number of students that
induce a conflict between courses. We therefore treat the weight of a
conflict $C_{i,j}$ as a random variable with a discrete support on
$\{0,1,\ldots, \min\{\nu_i,\nu_j\}\}$. Here $0$ is a valid value and
specifies that the conflict is not present. We call the vector $\nu$
and the matrix $C$ the uncertain parameter of our model. We call the
set of possible realizations of the parameters the uncertainty set
${\cal U}$ \cite[Def. 1.2.1]{ben2009robust}.  Given this uncertainty
in our model we have to revisit the hard constraints
\ref{enum:conflicting} and \ref{enum:cap} as they depend on these
random values.

\section{Robustness Measures}
\label{sec:robust}

As previously discussed, there are two possibilities for the time when
the scheduling is performed. The first variant is to start the
scheduling process after students have registered for all exams they
want to take. At that time the value for each random variable is
known, which gives us a model without uncertain parameters that can be
solved conventionally. The other variant, which is the focus in this
work, is to schedule the exams before registration takes place. This
presents us with challenges regarding several aspects of the
scheduling process. First and foremost we do not know how many
students attend an exam. Furthermore we do not know which exams are
actually in conflict.

As introduced in Section \ref{sec:uncertain_model} the curriculum
information provides us with upper bounds on the number of students
attending an exam and the data for which exams can be in conflict as
well as the upper bound on the number of students inducing the
conflict. However the range for the uncertain parameters is quite
large and therefore using their upper bounds is problematic as
this can and most of the time will lead to infeasible problems.

For optimization problems under uncertainty we can use the methods of
robust or stochastic optimization to find a solution that is feasible
for all or at least most realizations of the uncertain data.  In this
section we will give a brief overview of common robust and stochastic
optimization techniques for handling uncertainty in the problem
model. For an extended discussion on these topics we refer to Ben-Tal
et\,al. \cite{ben2009robust} for robust optimization and Birge and
Louveaux \cite{BL:2011} for stochastic optimization.

The basic optimization problem we want to optimize can be defined as
a general linear program.

\begin{align}
  \label{eq:opt_model}
  \min_x \quad &c^T x \\
  s.t. \quad &Ax \leq b \nonumber\\
               & x \geq 0 \nonumber
\end{align}

For a detailed description on how the examination timetabling problem
can be expressed as a linear program see \cite{McCollum2012}. In our
case the weights $c$, the constraint matrix $A$ and the vector $b$ are
subject to uncertainty. However the uncertainty in $c$ is ignored and
we focus on the uncertainty in $A$ and $b$ as these can lead to
violations of constraints in contrast to the impact of $c$ being local
to the objective value. Each uncertain variable $\zeta$ in the
uncertainty set ${\cal U}$ can be seen as
$\zeta = \bar \zeta + \delta$, where $\bar \zeta$ is the nominal value,
i.\,e., the value if no deviation from the expected scenario is
present, and the disturbance $\delta$ representing the possible
deviations from the expected scenario. Please note that $\delta$ can
be negative. A timetable is called nominal feasible if there is no
violation of the hard constraints when using the nominal values for
each uncertain parameter. We further call a timeslot feasible if there
is no violation of the hard constraints present for the exams in this
timeslot otherwise the timeslot is called \emph{disturbed}. Each
timeslot contains a subset of the exams $\cal E$ with an assignment of
rooms for each exam. We call such an assignment of rooms for an exam
$e$ room pattern $RP(e)$ with the corresponding capacity
$\kappa(RP(e)) = \sum_{r\in RP(e)} \kappa(r)$.

We will use this general model to introduce the different robustness
approaches and will discuss the implications of the uncertain
examination timetabling problem for the different approaches.

\subsection{Strict Robustness}
\label{sec:strict}

The first and simplest approach for robust optimization is called
strict robustness. A solution is called strict robust, iff it is
feasible for all realizations of the uncertain data. The tractability
of this approach is in large parts dependent on the uncertainty
set. In the case of examination timetabling we have an interval
uncertainty set ${\cal U}$. For interval uncertainty we get a linear
program with only linear overhead in the number of constraints. In our
particular case we can just use the upper bound on all uncertain
parameters. A solution for the resulting linear program will be
feasible for all possible scenarios of the uncertain data. The
existence of a feasible solution for the linear program given our
uncertainty set ${\cal U}$ based on curriculum data is highly unlikely
due to the far too pessimistic upper bounds on the number of students
and the pessimistic assumption that all possible conflicts are present. However
with more realistic bounds this approach could get viable, but better
information is usually not available.

\subsection{Stochastic Optimization}
\label{sec:stoch}

Another approach to handle uncertainty in the model is stochastic
optimization, see~\cite{BL:2011}. In contrast to robust optimization,
that tries to find the best feasible solution for all possible
scenarios of the uncertain parameters, stochastic optimization uses
stochastic measures to optimize the model.

\paragraph{Two-stage Stochastic Optimization}

In the two-stage approach we categorize the variables into two
distinct groups. The first group are the primary, \emph{first-stage}
variables. These variables have to be set before the scenario is
realized. The second group, the \emph{second-stage} variables however
can be set after the scenario is known. In this setting the linear
program transforms to

\begin{align}
  \label{eq:two-stage}
  \min_x \quad &c^T x + E[\min_y~ q^T y]\\
  s.t. \quad &Ax = b \nonumber\\
               & Tx + Wy = h \nonumber\\
               & x \geq 0 \nonumber \quad .
\end{align}
In this model the variables $x$ are the \emph{first-stage} and the
variables $y$ the \emph{second-stage} variables.  The goal of the
optimization is to find an optimal assignment of the
\emph{first-stage} variables, such that in expectation the objective
depending on the \emph{second-stage} variables is minimal.

In the case of examination timetabling a possible division of the
variables is to set the variables specifying the timeslot of an exam
as \emph{first-stage} and the variables specifying the rooms as
\emph{second-stage} variables. We however do not have the exact
distributions for our random variables and they might change over
time. Given this problem there are two approaches to still use the
technique of stochastic optimization. The first is to use some kind of
fitted distribution, e.\,g., normal distribution in the optimization
process in the hope that the actual numbers do not deviate too heavily
from the fitted distribution. The other approach is to calculate an
empirical distribution from the previous years. As with a fitted
distribution the usefulness depends heavily on the deviations of the
actual scenario.

\subsection{Recoverable Robustness}
\label{sec:recover}

In some cases we might have the situation that the
error situation is quite unlikely. In such a case using stochastic
optimization or especially strict robust optimization might be too
conservative. In this section we introduce the principle of
recoverable robustness. In recoverable robustness the scheduling uses
the nominal values for all uncertain parameters. Therefore the
returned solution is allowed to become infeasible when the scenario is
realized. However for each solution to be feasible in the recovery
robust model there needs to exist an algorithm to recover from the
infeasibility.

In theory recovery can become quite expensive in regard to the objective
value therefore in practice the cost of the recovery is often bounded. These bounds
come in two variants, namely with a fixed budget or as an
addition to the objective value to be minimized.

Given the linear program with the nominal variables $\bar A$ and
$\bar b$, i.\,e., the values if no errors are present compared to the
estimation, and the recovery matrix $\hat A$, i.\,e., the influence of
the recovery vector $y$ in the constraints, we get for a fixed budget
$D$ the problem
\begin{align}
  \label{eq:fixed_recovery}
  \min_x \quad &c^T x\\
  s.t. \quad &\bar Ax \leq \bar b \nonumber\\
  &\forall (A,b)\in {\cal U}~\exists y: Ax+\hat{A}y \leq b\nonumber\\
  &d^Ty \leq D\nonumber\\
  &x \geq 0\nonumber\quad
\end{align}
and the problem for a flexible budget $\lambda$ to be minimized
\begin{align}
  \label{eq:flex_recovery}
  \min_{x,\lambda} \quad &c^T x + t\lambda \\
  s.t. \quad &\bar Ax \leq \bar b \nonumber\\
  &\forall (A,b)\in {\cal U}~\exists y: Ax+\hat{A}y \leq b\nonumber\\
  &d^Ty \leq \lambda \nonumber\\
  &x \geq 0\nonumber\quad .
\end{align}

In academic timetabling it is possible, due to a change in the staff
or similar occasions, that a released timetable might become
infeasible, i.\,e., a lecturer might change to another university and
the replacement lecturer can not lecture the course or exam in the
assigned timeslot. In this case the afflicted course or exam needs to
be rescheduled, however this rescheduling should not influence the
whole timetable but be mostly local, e.\,g., should not affect more
than two timeslots.

In the case of data uncertainty of the number of students taking an
exam we can not limit our robustness approach to single disturbances
only. When using estimations for these numbers as nominal values in
reality we can expect to have a deviation from these values for each
exam. As such there can be several exams that have an insufficient
room assignment and our optimization approach needs to handle all
possible sets of these infeasible exams.

\paragraph{Our approach}
In our robust optimization approach on recovery we distinguish between
two types of variables similar to the two-stage stochastic
optimization. We consider the variables in our model specifying the
timeslot of an exam as \emph{first-stage} and the assignment of rooms
to exams as \emph{second-stage} variables. The reassignment of the
rooms of a timeslot are therefore the $y$ variables in
Eq.~\eqref{eq:fixed_recovery} and \eqref{eq:flex_recovery} with a cost
vector $\vec 0$. Only the timeslot has a direct impact on the quality of
the solution. This enables a potential recovery without an increase in
the quality of the solution. We first optimize the given instance
using an estimation for the different uncertain parameter. After the
actual registration of students for exams the scenario is realized,
which in turn can lead to timeslots that are infeasible in respect to
the actual values in the scenario. There are three types of exams in
such a timeslot. Exams with more, equal to or less actual students
than anticipated. Furthermore as we do not enforce the use of all
rooms in a timeslot we might have unassigned rooms available.

In the following we present two recovery strategies that are local to
the timeslot and therefore will not affect the solution quality.

\begin{algorithm}\caption{Heuristic Recovery\label{algo:heuristic_recovery}}
 \SetKwInOut{Input}{input}
 \SetKwInOut{Output}{output}
 \SetKw{in}{in}
\Input{Infeasible Timeslot $t$}
\Output{Feasible Assignment or notice of failure}
\For{exam $e$ \in $t$, $\phi(RP(e))) \geq \nu_e$}{
  Reduce rooms of exam e in t, such that the number of rooms is minimal
}

Assign free rooms to all exams $e$ in $t$ with $\phi(RP(e)) < \nu_e$\\
\eIf{feasible assignment of rooms for all exams in $t$ exists}{
return assignment}{
return failure}
\end{algorithm}

The goal of the \emph{Heuristic Recovery} strategy in
Algorithm~\ref{algo:heuristic_recovery} is to find a room assignment
for the infeasible timeslot without negatively affecting the feasible
exams.  If an exam has the same or fewer students than in the nominal
model, the recovery is only allowed to remove rooms but not to add
any. The strategy starts with removing all unnecessary rooms from
feasible exams. Afterwards all now unassigned rooms are assigned to
the infeasible exams. If such an assignment exists it is returned
otherwise the timeslot is marked as infeasible.  The advantage of this
strategy is that it only removes rooms from the exams that have an
feasible assignment and therefore the organizers of such an exam do not have to
increase their assigned staff for the exam.

\begin{algorithm}\caption{Complete Recovery\label{algo:complete-recovery}}
  \SetKwInOut{Input}{input}
  \SetKwInOut{Output}{output}

\Input{Infeasible Timeslot $t$}
\Output{Feasible Assignment or notice of failure}
Remove all rooms from the exams in $t$\\
Reassign all rooms to the exams in $t$
\end{algorithm}

The \emph{Complete Recovery} strategy in
Algorithm~\ref{algo:complete-recovery}, while still having no impact
on the objective value changes the room assignment completely. In the
first stage all room assignments are removed and afterwards a new
feasible assignment is computed for this timeslot. For the organizers
of an exam this means that they might need more persons supervising
the exam than originally planned even if their exam had a satisfying
room assignment. However if there is a possibility to recover from the
disturbances this strategy can find it. Both strategies do not
necessarily produce feasible timetables. If the strategy can not
recover all infeasible timeslots other recovery strategies have to be
used.

\subsection{Light Robustness}
\label{sec:light}

Light robustness is an approach that incorporates aspects of
stochastic and recoverable robustness, which was introduced by
Fischetti and Monaci \cite{Fischetti2009}.  The basic idea introduced
by Fischetti and Monaci is to first calculate the nominal optimal
solution and afterwards to find the most robust solution, where
objective values deviate only by a fixed amount.

Furthermore they introduce a heuristic approach for light
robustness. As before a nominal optimal solution is first calculated.
Afterwards a slack variable is introduced for each uncertain
constraint. Using these slack variables the model is solved again with
an upper bound on the difference to the nominal optimum. In this step
the new objective is to maximize the minimum of the normalized slack
variables. Unlike strict robustness the model does not guarantee the
feasibility in all scenarios of the uncertain data, but tries to find
the best solution in regards to a robustness measure within a given
range around the nominal optimum.  In a third optimization step a
solution is calculated that balances the normalized slack variables
while the objective value is only allowed to deviate by a fixed
value. In this step the minimum of the normalized slack variables as
calculated in the second optimization step is used as a lower bound
for the normalized slack variables.

In \cite{BW:18} we introduced an approach called probabilistic
curriculum-based examination timetabling (PCBETT) that is quite
similar to the heuristic approach of Fischetti and Monaci. The main
difference is that instead of first calculating the nominal optimum
and afterwards optimizing a robustness measure we incorporate the
robustness measure in regard to the number of students per exam directly into
the optimization. For this purpose we introduced the new soft
criterion \eqref{eq:old_pcb} in \cite{BW:18}.
\begin{equation}
  \max \left\{\frac{\nu(e)}{\kappa(RP(e))} \mid e \in \mathcal{E}\right.,
  \text{$RP(e)$ is room pattern of $e$ in the current
  timetable}\left.\vphantom{\frac{{\rm E}[\nu(e)]}{\kappa(RP(e))}}\right\}\label{eq:old_pcb}
\end{equation}
with $\kappa(RP(e)) = \sum_{r\in RP(e)} \kappa(r)$ the canonical
extension as introduced in the model of Sect.~\ref{sec:model}.  This
new soft criterion should be minimized.  As this introduced soft
criterion is non-linear we use a slight but
semantically equivalent reformulation for this paper that can be linearized by
standard transformations. 
\begin{enumerate}[label=(S\arabic*),labelindent=3pt,leftmargin=*]
\setcounter{enumi}{3}
\item $\displaystyle\min \left\{\frac{\kappa(RP(e))}{\nu(e)} \mid e \in
    \mathcal{E}\right., \text{$RP(e)$ is room pattern of $e$ in current timetable}\left.\vphantom{\frac{\nu(e)}{\kappa(p(e))}}\right\}$\label{enum:pcb}
\end{enumerate}
This reformulated soft criterion has to be maximized. In contrast to
the approach introduced by Fischetti and Monaci we use the soft
criterion directly in our optimization and only solve the problem once as
instances for the ETTP can be quite large. Note that this approach does not handle the uncertainty regarding conflicts.

\section{Instance Generation and Optimization}
\label{sec:framework}

In Sect. \ref{sec:robust} we introduced several strategies to get a
robustness aspect into the examination timetabling optimization. To
get an understanding of the performance of these different strategies
we compare them on two real world instances from the University
of Erlangen-N\"urnberg. However to increase our test size we introduce
an instance generation algorithm in Sect.~\ref{sec:instance} to generate random instances with
different characteristics in respect to structure of the conflicts. These instances were
than optimized by a simulated annealing algorithm described in
Sect. \ref{sec:opt}.

\subsection{Instance Generation}
\label{sec:instance}

In optimization it is often interesting to study real world instances
and the influence of different approaches on them. However to get a
deeper understanding of these techniques and to rule out unique
influences present in our real world instances we have to increase and
diversify the test data. There are two approaches to accomplish this
diversity. The first and often used is the collection of different
instances from other universities. In the case of the characteristics
of our model, e.\,g., the assignment of multiple rooms to an exam,
there exists to our knowledge no such collection. Furthermore it would
still be limited to only a small number of different instances.  To
have a diverse set of instances, where we can produce different
characteristics and study their influence on the performance of the
robustness approaches we introduce a random instance generation
framework.

\begin{algorithm}\caption{Instance Generation \label{algo:instance-generation}}
  \SetKwInOut{Input}{input}
  \SetKwInOut{Output}{output}
  \SetKw{to}{to}
  \SetKw{from}{from}
  \SetKw{in}{in}
  
  \Input{Number of exams $E$, Number of rooms $R$, Number of timeslots $T$}
  \Output{ETTP Instance}

  ${\cal R}$ := set of rooms of size $R$\\
  \For{$r$ \in ${\cal R}$}{
    \tcc{$v\sim{\rm NB}(0.2,25)$ in our tests, other distributions are possible}
    choose $v \sim {\rm NB}(0.2,25)$\\
    set $\kappa(r) := v$ 
  }
  ${\cal E}$ := set of exams of size $E$\\
  Generate random partition $EP=(EP_1,EP_2,\ldots,EP_T)$ of ${\cal E}$ into $T$ sets each of size between $1$ and $R$\\
  \For{$t$ \from $1$ \to $T$}{
    Generate random assignment $RP$ of rooms in ${\cal R}$ to exams in $EP_t$, which is a partition of ${\cal R}$ with sets of size at least $1$\\
    \For{$e$ \in $EP_t$}{
      \tcc{$n \sim{\rm Beta}(10,5)$ in our tests, other distributions are possible}
      choose $n \sim {\rm Beta}(10,5)$ \\
      $\nu(e)$ := $1 + \lfloor \kappa(RP(e)) \cdot n \rfloor$
    }
  }

  \For{$i$ \from $1$ \to $E$}{
    \For{$j$ \from $1$ \to $E$, $i\neq j$}{
      choose $u$ u.a.r. from $[0,1]$\\
      \tcc{$c(i,j)$ is the predefined probability for exams $i$ and $j$ to be in conflict}
      \eIf{$u < c(i,j)$}{
        $C_{i,j}$ := 1
      }{
        $C_{i,j}$ := 0
      }
    }
  }
\end{algorithm}

Algorithm~\ref{algo:instance-generation} shows the framework for our
instance generation which is highly flexible as several different
distributions that can be used for the random variables. We start with
generating a number of rooms. For our tests room sizes are sampled
according to a negative binomial distribution with the parameter
$p=0.2$ and $k=25$. This gives us a selection of rooms with mostly
medium to small rooms and only a few large rooms. Afterwards we
generate a random partition of the exams into $T$ sets, where $T$ is
the number of timeslots and each set of the partition contains between
$1$ and up to $|\cal R|$ exams. For each timeslot we generate a random
assignment of rooms to the exams in the timeslot, where a room is
assigned to exactly one exam. The number of students is then drawn at
random between $1$ and the assigned room capacity $\phi(RP(e))$. Using
this generation process we can ensure that without conflicts the
instance is feasible. Afterwards we generate conflicts for the exams
by predefined probabilities for each conflict. In our experiments we
used two different types of choices for the probability of the
conflicts, see Sect.~\ref{sec:experiments}.

\subsection{Optimization Framework}
\label{sec:opt}

As our models can get quite large and therefore can most often not
be solved to optimality we use a heuristic approach to solve the
examination timetabling problem. To perform the optimization of our
two real world instances and the instances generated by our instance
generation framework we implemented a simulated annealing
algorithm. The used algorithm is quite similar to the simulated
annealing algorithm described by M\"uhlenthaler in \cite{M:15}. It
uses the Kempe-Exchange neighborhood to select a new neighbor.

The differences between the simulated annealing algorithm as
introduced in~\cite{M:15} and the algorithm used for our optimization
are the feasibility of the room assignment and the cooling scheme.

A neighbor is only considered if there exists a room assignment for
each of the two changed timeslots. We used Cplex \cite{Cplex:2014} to
calculate the room assignments for each timeslot. This is necessary to
ensure the feasibility of each consecutive step in the annealing
phase. For the nominal solution without the use of a robustness
measure we just return the first solution found by the solver. When
using the PCBETT approach the room assignment is instead optimized
using only soft constraint \ref{enum:pcb} as the objective and the
optimal solution of the room assignment in regards to \ref{enum:pcb} is
returned. As only one timeslot is considered in this optimization the
feasibility of the timeslot depends only on the hard constraint
\ref{enum:cap} which is satisfied by using soft constraint
\ref{enum:pcb} if a feasible solution exists.

Experiments performed using this simulated annealing algorithm showed
a fast convergence to a local minimum with only a small chance to
leave this local optimum. To counter this phenomenon we modified the
cooling scheme. After a predefined amount of iterations we reset the
heat level. Furthermore to counter a strong drift while not in a local
optimum we doubled the cooling rate if a better solution was accepted
in the iteration.

Using this cooling scheme we could increase the performance of the
simulated annealing algorithm, however the problem could not be
avoided. To further improve the returned solution we started the
simulated annealing algorithm several times but with fewer
iterations. The best solution found in the different runs was then
returned, which could significantly improve the performance of the
overall algorithm.

As a starting point we used a simple heuristic that schedules the
exams in decreasing order in respect to the number of students taking
the exam. Rooms are assigned in decreasing order in respect to the
capacity of the room until the assigned capacity is larger than the
number of students taking the exam. If not enough rooms are available
in the timeslot the next timeslot is used. In practice the heuristic
proved to be sufficient for an initial feasible solution. The
objective value however was quite far from optimal, but can be
optimized by the simulated annealing quite drastically.

\section{Robustness Evaluation}
\label{sec:experiments}

To compare the different robustness approaches introduced in
Sect~\ref{sec:robust} we used two real world instances from the University of Erlangen-N\"urnberg and the instance generation framework
introduced in Sect. \ref{sec:instance}. With this framework we generated $20$ instances
for the $5$ different instance generation settings as shown in Table~\ref{tab:setting} to have a
better understanding of the approaches for different characteristics
in examination timetabling instances.

\begin{table}
  \centering
  \setlength{\tabcolsep}{2.5pt}
  \begin{tabular}{| c | c | c | c | c | c | c |}
    \hline
    Scenario & $|\cal E|$ & $|\cal R|$ & $T$ & $\nu_e$ & $\kappa$ & Conflict generation \\
    \hline
    Scenario 1 & 120 & 8 & 30 & ${\rm Beta}(10,5)$ & ${\rm NB}(0.2,25)$ & chosen u.a.r. with $p=0.1$ \\
    \hline
    Scenario 2 & 120 & 8 & 30 & ${\rm Beta}(10,5)$ & ${\rm NB}(0.2,25)$ & chosen u.a.r. with $p=0.25$ \\
    \hline
    Scenario 3 & 90 & 8 & 30 & ${\rm Beta}(10,5)$ & ${\rm NB}(0.2,25)$ &  Poisson-binomial with  $p_e \sim {\rm Beta}(4,6)$ \\
    \hline
    Scenario 4 & 120 & 8 & 30 & ${\rm Beta}(10,5)$ & ${\rm NB}(0.2,25)$ &  Poisson-binomial with  $p_e \sim {\rm Beta}(4,6)$ \\
    \hline
    Scenario 5 & 150 & 8 & 30 & ${\rm Beta}(10,5)$ & ${\rm NB}(0.2,25)$ &  Poisson-binomial with  $p_e \sim {\rm Beta}(4,6)$ \\
    \hline
    Summer 2018 & 405 & 13 & 75 & \multicolumn{3}{|c|}{Summer term 2018} \\
    \hline 
    Winter 2017 & 411 & 13 & 75 & \multicolumn{3}{|c|}{Winter term 2017} \\
    \hline
  \end{tabular}
  \caption{Instance generation settings used in the evaluation of the robustness approaches}
  \label{tab:setting}
\end{table}

In our test sets we used two different types of choices for the
probability of a conflict. The first and most simple conflict
generation generates a conflict between two exams uniform at random
(u.a.r.) with a given probability. As such exams have around the same
number of conflicts each. This however is quite far from an actual
distribution of conflicts in real word data. To increase the
similarity between the randomly generated instances and actual real
word instances we generated the conflicts Poisson-binomial
distributed. For each exam we draw a conflict probability according to
a ${\rm Beta}(4,6)$ distribution. This beta distribution creates the
situation that most exams have a medium to low number of conflicts
with only a few exams having a high number of conflicts. The
probability of a conflict is then calculated as the product of the
exam probabilities of the exams in the conflict. This results in a
Poisson-binomial distribution for the conflicts. The number of exams
and timeslots are chosen to have a ratio similar to our real world
instances.

As described in \cite{BW:18} for our real world instances we used the
information taken from the previous years for the number of students
taking an exam and the number of students inducing a conflict is the
arithmetic mean taken over the previous years.

Each generated instance is then optimized by the simulated annealing
algorithm introduced in Sect.~\ref{sec:opt} with parameters
$t\in[1,100]$ as heat level, a cooling limit of $5\,000$, $50\,000$
iterations and $8$ repetitions. We used the soft constraint weights
$300$ for \ref{enum:two-in-a-row}, $150$ for \ref{enum:two-in-a-day}
and $5$ for \ref{enum:period-spread} with a period spread of
$\lambda=4$.  Each instance is optimized twice, once with the soft
constraint \ref{enum:pcb} and weight $3000$ and once without it. We
then take the arithmetic mean over the $20$ different instances for
each soft criterion. As already a single optimization run of a single
instance has a runtime up to an hour it is not possible to increase
the test size significantly and still evaluate all tests in a
reasonable amount of time. For the two real world data instances we
increased the iterations to $150\,000$ and the cooling limit to
$50\,000$.

\begin{table}
  \centering
    \setlength{\tabcolsep}{3pt}
  \begin{tabular}{| c | c | c | c | c | c | c | c | c | c |}
    \hline
    & \multicolumn{4}{|c|}{Nominal} & \multicolumn{5}{|c|}{PCBETT}\\
    \hline
    Scenario & Objective & \ref{enum:two-in-a-row} & \ref{enum:two-in-a-day} & \ref{enum:period-spread} & Objective & \ref{enum:two-in-a-row} & \ref{enum:two-in-a-day} & \ref{enum:period-spread} & \ref{enum:pcb}\\
    \hline &&&&&&&&& \\[-1em]
    \hline 
    Scenario 1 & 15 & 0 & 0 & 3 & 294 & 0 & 0 & 58.9 & 1.62114\\
    \hline
    Scenario 2 & 38649.75 & 2.9 & 57.75 & 5823.45 & 39645.25 & 2.8 &62.95 & 5872.55 & 1.08674 \\
    \hline
    Scenario 3 & 56.25 & 0 & 0 & 11.25 & 205.75 & 0 & 0 & 41.15 & 1.62793\\
    \hline
    Scenario 4 & 7383.25 & 0.2 & 1.55 & 1418.15 & 8397 & 0.15 & 2 & 1610.4 & 1.27832\\
    \hline
    Scenario 5 & 24671.5 & 1.45 & 28.8 & 3983.3 & 25247.5 & 1.6 & 26.5 & 4158.5 & 1.12212\\
    \hline
    Summer 2018 & 8645 & 0 & 33 & 739 & 9170 & 0 & 38 & 694 & 2.21794\\
    \hline
    Winter 2017 & 13925 & 0 & 57 & 1075 & 14255 & 2 & 53 & 1141 & 1.92650 \\
    \hline
  \end{tabular}
  \caption{Influence of the pcbett approach on the objective function. Each value is the arithmetic mean over 20 generated instances 
    generated per scenario as well as the objective function values for the two real world data instances.}
  \label{tab:obj}
\end{table}

\begin{figure}
  \centering
  \input{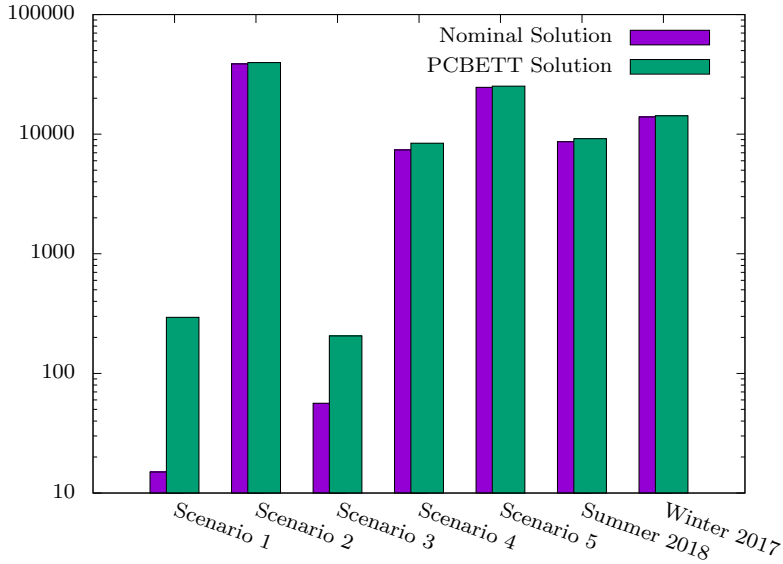}
  \caption{Arithmetic mean of the weighted objective values calculated over 20 generated instances per scenario and the weighted objective values of the two real world data instances.}
  \label{fig:weighted_sums}
\end{figure}
The results for the experiments are presented in Table~\ref{tab:obj}
and the weighted sum of the objective values is additionally
visualized in Fig.~\ref{fig:weighted_sums}. Note that the weighted sum
is only calculated over the soft criteria \ref{enum:two-in-a-row} to
\ref{enum:period-spread} as the nominal solution does not use soft
criterion \ref{enum:pcb}. The value of \ref{enum:pcb} in the nominal solutions is therefore very close to $1$. Through the addition of the soft criterion
\ref{enum:pcb} the objective values of the soft criteria
\ref{enum:two-in-a-row} to \ref{enum:period-spread} tend to be worse,
however the difference is quite low.

The second test uses the instances and timetables generated by the
first test and introduces errors into the timetables. As argued in
Sect.~\ref{sec:uncertain_model} each exam is subject to errors. The
new value for the number of students for exam $e$ is chosen as
$\nu_{e,new} \sim N(\nu_{e,old}, \frac{\nu_{e,old}}{5})$. For each
instance this experiment is repeated $100$ times and the timetables
are evaluated with the new values $\nu_e$. For each timetable the
number of infeasible timeslots are counted and the arithmetic mean is
calculated.

\begin{table}
  \centering
      \setlength{\tabcolsep}{4pt}
  \begin{tabular}{| c | c | c | c | c | c | c |}
    \hline
    & \multicolumn{3}{|c|}{Nominal} & \multicolumn{3}{|c|}{PCBETT}\\
    \hline
    Scenario & Unmodified & Heuristic & Complete & unmodified & Heuristic & Complete \\
    \hline &&&&&& \\[-1em]
    \hline 
    Scenario 1 & 8.05500 & 4.46500 & 1.34250 & 0.03150 & 0.01750 & 0.00100\\
    \hline
    Scenario 2 & 8.57250 & 5.44850 &  1.85000 & 4.01350 & 3.46500 & 1.49700 \\
    \hline
    Scenario 3 & 6.57750 & 2.84650 & 1.02100& 0.16700 & 0.09750 & 0.03100\\
    \hline
    Scenario 4 & 8.29000 & 5.19900 & 2.00100 & 1.42950 & 1.07250 & 0.33550 \\
    \hline 
    Scenario 5 & 10.53950 & 7.43650 & 2.30950 & 4.03200 & 3.51250 & 1.38050 \\
    \hline
    Summer 2018 & 13.18000 & 3.13000 & 0.65000 & 0 & 0 & 0 \\
    \hline
    Winter 2017 & 12.58000 & 2.69000 & 0.16000 & 0 & 0 & 0 \\
    \hline
  \end{tabular}
  \caption{Number of disturbed timeslots before and after robust recovery for the nominal and the pcbett solutions.}
  \label{tab:disturbed}
\end{table}

Table~\ref{tab:disturbed} shows the results for the arithmetic mean
over the $20$ test instances and the $100$ disturbances. We can see,
that the number of disturbed timeslots when using the soft criterion
\ref{enum:pcb} is significantly lower than the number of disturbed
timeslots for the unmodified nominal solution. It is also noticeable
that the repair heuristic and especially the complete recalculation
can significantly reduce the number of timeslots that are
infeasible. For the heuristic the number is however still bigger than
the unmodified optimization results with soft criterion
\ref{enum:pcb}. If we allow the complete recalculation of the
disturbed timeslots the actual infeasible timeslots can be reduced
significantly while still no change in the objective function
occurs. However as discussed in Sect.~\ref{sec:recover} this
recalculation can change the assignment of rooms quite drastically and
therefore have negative impact on the acceptance of the teaching
staff.

\section{Conclusion}
\label{sec:conc}

In this work we introduced several approaches of robust and stochastic
optimization and discussed their applicability for the examination
timetabling problem. We furthermore introduced a new random instance
generation algorithm, which allowed the generation of an increased
number of test instances with different characteristics. Given the
quite large uncertainty sets resulting from curriculum information we
discussed the problems for the different robustness approaches. We
introduced two possible recovery algorithms that do not change the
function value as they only consider one timeslot. The experimental
results showed that especially the complete recovery algorithm was
able to repair most of the disturbed timeslots however at the cost of
the complete reassignment of rooms which might have a negative impact
on the acceptance from the staff holding the exams. As a light
robustness approach we could experimentally show that the introduction
of another soft constraint was able to significantly reduce the number
of timeslots that get disturbed, i.\,e., have exams with an
insufficient room assignment, with only a small impact on the overall
quality of the solution.

\bibliographystyle{plain}
\bibliography{literature}

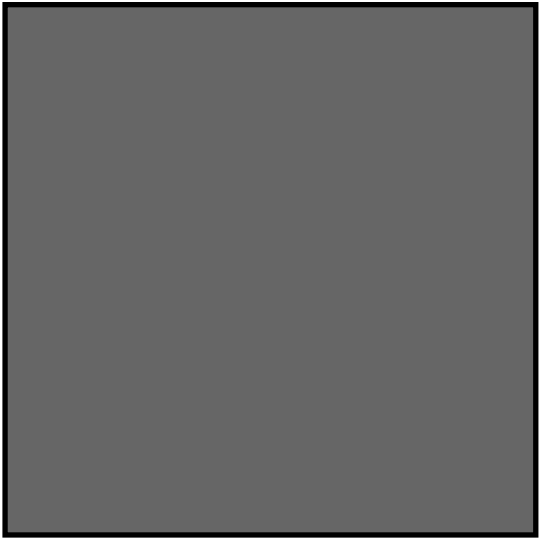
\end{document}